# (KCC2017 우수논문) 깊은 신경망에서 단일 중간층 연결을 통한 물체 분할 능력의 심층적 분석†
# (Investigating the feature collection for semantic segmentation via single skip connection)

임 종 화 ††
(Jonghwa Yim)

손 경 아 †††
(Kyung-Ah Sohn)

**요 약** 최근 심층 컨볼루션 신경망을 활용한 이미지 분할과 물체 위치감지 연구가 활발히 진행되고 있다. 특히 네트워크의 최상위 단에서 추출된 특징 지도뿐만 아니라, 중간 은닉 층들에서 추출된 특징 지도를 활용하면 더욱 정확한 물체 감지를 수행할 수 있고 이에 대한 연구 또한 활발하게 진행되고 있다. 이에 밝혀진 경험적 특성 중 하나로 중간 은닉 층마다 추출되는 특징 지도는 각기 다른 특성을 가지고 있다는 것이다. 그러나 모델이 깊어질수록 가능한 중간 연결과 이용할 수 있는 중간 층 특징 지도가 많아 지는 반면, 어떠한 중간 층 연결이 물체 분할에 더욱 효과적일지에 대한 연구는 미비한 상황이다. 또한 중간층 연결 방식 및 중간층의 특징 지도에 대한 정확한 분석 또한 부족한 상황이다. 따라서 본 연구에서 최신 깊은 신경망에서 중간층 연결의 특성을 파악하고, 어떠한 중간 층 연결이 물체 감지에 최적의 성능을 보이는지, 그리고 중간 층 연결마다 특징은 어떠한지 밝혀 내고자 한다. 그리고 이전 방식에 비해 더 깊은 신경망을 활용하는 물체 분할의 방법과 중간 연결의 방향을 제시한다.

**키워드** : 물체 분할, 특징 지도, 중간층 연결, 깊은 컨볼루션 신경망, 중간층

**Abstract** Since the study of deep convolutional neural network became prevalent, one of the important discoveries is that a feature map from a convolutional network can be extracted before going into the fully connected layer and can be used as a saliency map for object detection. Furthermore, the model can use features from each different layer for accurate object detection: the features from different layers can have different properties. As the model goes deeper, it has many latent skip connections and feature maps to elaborate object detection. Although there are many intermediate layers that we can use for semantic segmentation through skip connection, still the characteristics of each skip connection and the best skip connection for this task are uncertain. Therefore, in this study, we exhaustively research skip connections of state-of-the-art deep convolutional networks and investigate the characteristics of the features from each intermediate layer. In addition, this study would suggest how to use a recent deep neural network model for semantic segmentation and it would therefore become a cornerstone for later studies with the state-of-the-art network models.

**Key words** : Semantic segmentation, Feature map, Skip connection, Deep convolutional network, Intermediate layer

## 1. Introduction

One of the important discovery of convolutional neural network is that the features after convolutional operation comprise both semantic and spatial information. The early study of semantic segmentation using deep neural network includes fully convolutional architecture. For instance, the study [1] designed fully convolutional network for pixel-wise semantic segmentation, inspiring us how to use features after multiple convolutional operations. This is meaningful in that the end-to-end classifier, convolutional neural network, can resolve object segmentation, an insoluble problem beforehand. Previously, many approaches were not an end-to-end, but a combination of convolutional network and traditional image processing technics. The first study of object localization using such combination [2] broke the records in object localization. Since then, convolutional network became a prospective method for object detection. In the meantime, this study [2] and consequent studies [3], [4] were not a pixel-wise segmentation though they took advantages of convolutional architecture. The study [4] was an end-to-end model by fully utilizing the feature map from convolutional network to generate candidates for precise object localization. This method is similar to the first pixel-wise segmentation neural network model [1] in that both models use features maps from fully convolutional structure to attain semantic and spatial information.

Meanwhile, the concept of skip connection was firstly introduced in the previous studies [1], [5]. The idea behind skip connection is that the features from intermediate layers contain more spatial information than the one from the uppermost layer, allowing us to aggregate the features from intermediate layers and merge onto the features from the final layer. The advantage of this is that we can elaborate the result of semantic segmentation by utilizing both the features of lower layers, which are localized and less downsampled, and that of upper layers that contains semantically meaningful blobs. Hence, skip connection became widely popular and many researches use it to obtain even better accuracy. For example, the study [6] leverages all possible intermediate features at all hidden layers to generate top-down or bottom-up feature pyramids, and claims the state-of-the-art performance. However, these pyramidal operations require tremendous time and computational effort. Moreover, some intermediate features might be inefficient for object segmentation. Therefore, we need comprehensive research into the characteristics of the features from many possible intermediate layers and its skip connections as well.

Until now, many semantic segmentation studies exploit VGG network model [7], which is relatively shallow than recent models [8], [9]. The main reason is that it requires relatively low computational memory and maintains local information of the object in an image relatively well. However, many recent models comprise deeper structure than VGG network and achieve better results in image classification task in popular competitions such as ILSVRC [10]. Considering that VGG network model is old in the history of deep neural network, we need to apply the cutting-edge deep neural network models for semantic segmentation.

## 2. Related Work

Since the study [1] inspired us in many ways for semantic segmentation using convolutional neural network, there came out many branches of approaches of semantic segmentation. To restore the encoded data and upsample the feature map, we typically stack transposed convolution (or deconvolution) layer onto the original model to generate original-size feature map and achieve pixel-wise prediction. A well-known example of the use of deconvolution via upsampling operation is a Segnet [11]. This paper showed that a symmetric structure of upsampling and transposed convolution attains better accuracy than the previous method [1]. For better and accurate restoration while upsampling the feature map, some researches [12] use hole algorithm (or dilated convolution). A characteristic of architecture with this convolution is that it does not contain downsampling or upsampling operations. Therefore, the advantage of this structure is that it preserves details while maintaining the original receptive field of transposed convolutional structure, thereby achieving a better accuracy than transposed convolution.

Some of recent methods applied skip connection broadly to use low-level feature maps and achieved the state-of-the-art performance. These approaches include SharpMask [13] which is an advanced model of DeepMask [14] through skip connection for mask segmentation, Stacked Hourglass networks [15] for human body segmentation by residual modules, and Recombinator network [16] for face detection, all of which associate features maps from low level layers with the feature maps from fully convolutional network for pixel-wise prediction. Some researches arbitrarily determine which intermediate layer they would use without further analysis, or other researches comprehensively include and merge all hidden layers. These approaches seem inefficient and give insufficient reason to do so.

Although the majority of the recent studies of semantic segmentation adapted VGGNet model as stated earlier, some recent studies [17], [18] include research exploiting deep neural network models such as Resnet-101 or Inception network. Furthermore, [17] exploit dilated convolution instead of usual downsizing layer and transposed convolution layer to solve the problem with low resolution feature map.

## 3. Purpose of Study

As stated earlier, to improve semantic segmentation, many studies aggregate the features of intermediate layers and merge onto the features from the final layer. This can be achieved by skip connection. However, we still do not have enough knowledge on choosing the efficient skip connection and explaining the characteristics of each intermediate feature. Therefore, in this study, we perform an exhaustive research on the features from each layer and many possible skip connections in cutting-edge deep neural network models to suggest better skip connections among candidates.

## 4. Method for Experiment

### 4.1. Models

Previous researches, as stated in earlier chapters, majorly utilize VGGNet [7] for semantic segmentation. GoogLeNet [8] model is also used in a few researches, but does not show significant improvement and both are old models now. To keep up with recent studies in neural network, in this study, we use an up-to-date deep neural network model, Inception-v3 [19] to get aligned with the state-of-the-art performance on image classification and to scrutinize the characteristics of various features from many different levels of layers.

### 4.2. Skip Connections

There are several possible skip connections to intermediate layers. If we try to make use of all latent skip connections, it might require enormous amount of computing capacity. Moreover, some of them are thought to have nearly identical feature maps with each other when they are adjacent and have only few convolutions between them. When an image resolution is reduced or after image going through several convolutional operations, we might want to verify the characteristics and performances of skip connection. Therefore, among many possible intermediate features we can choose in Inception-v3 architecture, we select seven intermediate layers for skip connections and semantic segmentation. Each different intermediate feature map has different resolution from each other. Therefore, we need to upsample them before merging with the uppermost layer's feature map. During the experiment, we discovered that the feature map, without any further convolutional operation, could not properly localize objects. That is why we need a $n \times n$ extra convolutional layer to generate the intermediate feature map before merging. Then we apply a $1 \times 1$ convolutional operation to generate class-wise feature maps. The size of extra convolution filter is $n$ and is chosen empirically. If $1 \times 1$ convolution, without $n \times n$ convolution, is solely applied, then all the possibilities of 21 classes are abnormally high at all regions. This is because, without expanding the region of interest as assumed as an object size, the model cannot properly distinguish what each block of feature map might be. Therefore, we apply either $3 \times 3$, $5 \times 5$, or $7 \times 7$ convolution on the feature map from intermediate layer. This process is shown in Fig. 1.

### 4.3. Upsampling

After obtaining the feature maps from the final layer and intermediate layers, we upsample the feature maps for semantic segmentation. Although many experiments perform pixel-wise prediction, in this study, we perform box-wise prediction, dividing the image into $67 \times 67$ areas. This experimental setting eases the device memory and time requirement, letting our experiment easier. The architecture of Inception-v3 based model in our experiment is shown in Fig. 2 and an overview of upsampling and merging the feature maps is displayed in Fig. 3. In Fig. 2, numbering means seven skip connections in our experiment as stated in chapter 4.2. This numbering is used to refer each skip connection throughout the later chapters in this paper.

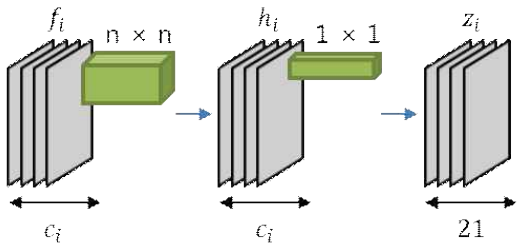

**Fig. 1. Feature map implementation of our model. The features $f$ at the $i^{th}$ layer is extracted and the convolution of $n \times n$ is applied to obtain intermediate feature $h$. Then, convolution is applied to generate feature map $z$ with respect to classes.**

## 5. Experiment

### 5.1. Datasets and Metrics

Semantic segmentation is separating target object out of background in an image. The popular datasets for this task are PASCALVOC [20] and COCO detection challenge [21] datasets. Although the most prevalent dataset is MSCOCO cls-loc dataset, for easier experiment and comparison, in this study, we train and measure the performance of object segmentation using

PASCALVOC 2012 train and validation datasets each. Unlike image classification, this task needs to find out exact location and outer shape of target object given a 2D image. Therefore, there are a few unique metrics to measure the performance of this task.

$$Pixel\ accuracy : \sum_i \frac{n_{ii}}{\sum_i t_i} \quad ..............................(1)$$

$$mean\ accuracy : \left(\frac{1}{n_d}\right) \sum_i \frac{n_{ii}}{t_i} \quad .........................(2)$$

$$mean\ IoU : \left(\frac{1}{n_d}\right) \sum_i \frac{n_{ii}}{\left(t_i + \sum_j n_{ji} - n_{ii}\right)} \quad ...........(3)$$

In the formulas, $n_{ii}$ is the number of pixels of correct prediction on $i^{th}$ class while $n_{ji}$ is the number of predictions of $j^{th}$ class on pixels belongs to $i^{th}$ class, and $t_i$ is the number of all pixels belongs to $i^{th}$ class. In essence, mean IoU (intersection over union) metric is preferred since they can cover the total number of correct predictions over mistaken predictions. Pixel accuracy can still be high even though the prediction indicates almost all background. Thus, in this study, mean IoU metric is used to measure the performance of each skip connections.

### 5.2. Evaluation

After performing exhaustive experiments, we verify the most effective skip connection in Inception-v3 model. We train the model for around 100 epochs on PASCALVOC training dataset. The results are shown in Table 1. Then we modify the filter size of intermediate layers. The results are shown in Table 2.

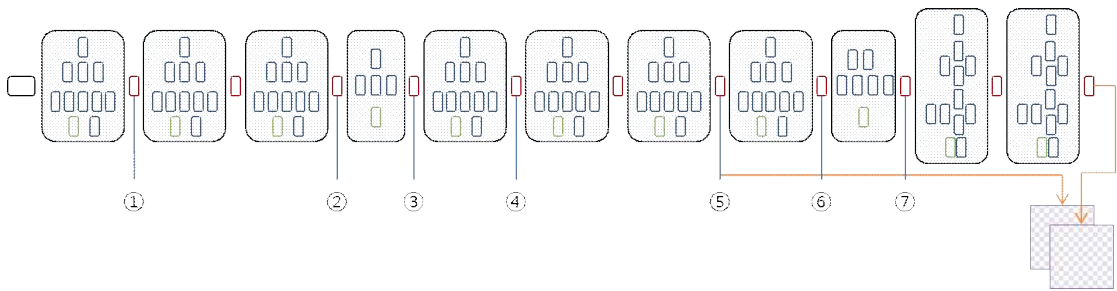

**Fig. 2. The structure of Inception-v3 model and its candidates of skip connections. We choose seven latent skip connection among many candidates. We number each skip connection for our experiment. Then the feature map of intermediate layers is summed to that of the final layer.**

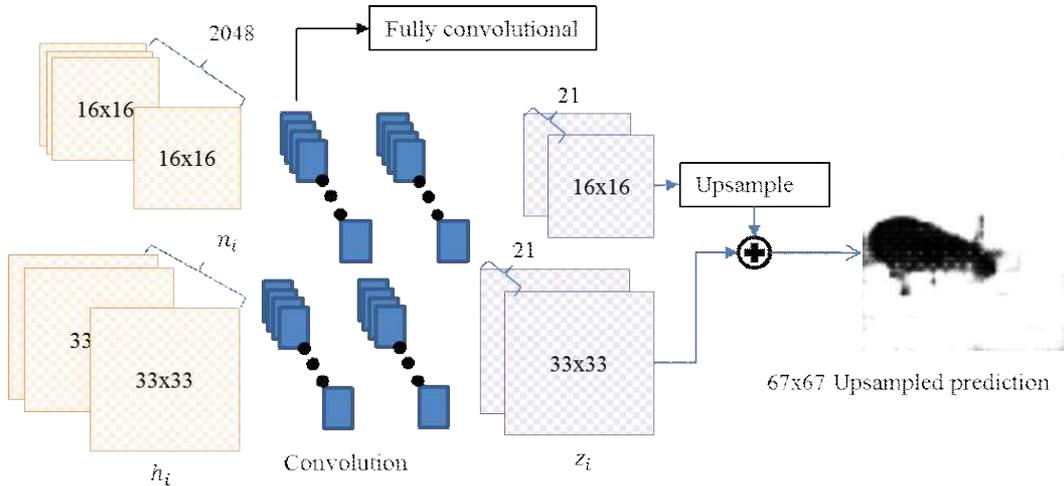

**Fig. 3. The overview of upsampling and merging two feature maps from different layers in Inception-v3 architecture. After being applied convolution twice, feature map $h_i$ from an $i^{th}$ intermediate layer generates the feature map $z_i$, and then is upsampled to generate 67 × 67 prediction map.**

Estimated size of object features that we attain from intermediate layers may vary. Hence, we modify and test different convolutional filter size to obtain intermediate feature $h$. The result in Table 2 gives an idea of choosing the filter size.

**Table 1. Result comparison. In this table, numbering matches the numbers used in Fig. 2. It indicates the best skip connection among seven possible connections.**

| Connections | mIoU |
|---|---|
| No connection | 30.06 |
| 7 | 30.48 |
| 6 | **39.92** |
| 5 | **33.67** |
| 4 | 31.57 |
| 3 | 29.81 |
| 2 | 28.29 |
| 1 | 28.9 |

**Table 2. mIoU result by filter size of intermediate layers. The size of feature maps from layers marked number 5 and 6 is 33 × 33. Incrementing filter size improves the result of number 6 skip connection.**

| Connections | filter size 7 | filter size 5 |
|---|---|---|
| 6 | **39.92** | 32.33 |
| 5 | 31.84 | 33.67 |

As stated in previous section 4.1, the state-of-the-art methods for classification task use deep convolutional networks such as Inception-v3. On the contrary, most of the previous methods for segmentation task still utilize VGG16 network model. In our experiment, semantic segmentation using Inception-v3 with skip connection shows decent background masking segmentation. However, our model was not comparable to the famous FCN method in terms of mean IoU. We compare our model with the popular methods [1], [5], [22] in Table 3.

**Table 3. Comparison with other methods.**

| | Background | Mean IoU |
|---|---|---|
| Inception-v3 | 89.7 | 39.9 |
| Hypercolumn [5] | 88.9 | 59.2 |
| DeconvNet [22] | 92.7 | 69.6 |
| FCN8s [1] | 91.2 | 62.2 |

Although this study is not aimed at the state-of-the-art performance, its thorough research on skip connection would be a cornerstone for successive researches on neural network deeper than VGG network model [7] and the usage of skip connection for semantic segmentation. Since it shows decent background masking precision, we would obtain better result in later study if we apply data augmentation and adapt additional technics such as conditional random field or dilated convolution.

## 5.3. Analysis of the Feature Maps

The experiment reveals that the features of upper layers are most effective for semantic segmentation. Features of top layer, without skip connection, contains class-wise information to classify the image, but rarely contains spatial information of each object, moreover, is a low-resolution map. However, intermediate layer of upper part of the network contains both semantic and localized information. The important thing is that the feature of second top layer, marked as "7" in our experiment, has little spatial information, giving no better result. This is because the size of the feature map is same as that of top layer, having the receptive field too large and loosing spatial evidence much. On the other hand, the features from bottom layers contain basic components of object such as edges or blobs. In Fig. 4 to 6, we visualize the output of each skip connection and make a comparison to analyze the differences.

In Fig. 4, 5, and 6, images with the first number "0" in file name contain airplanes and with "1" are training images. Second number in file name is a class label including a background label 20. In Fig. 4, the output feature map clearly visualizes the object semantically and spatially. However, in Fig. 5 and 6, the output feature maps, although they are still masking out a background area well, classify target object inaccurately. Therefore, many classes' feature maps wrongly respond to target object. Actually, this happens because the feature maps from lower layers contain spatial but not semantic information. Therefore, they tend to respond to objects' basic components. In Fig. 6, images of "1_0", "1_13", and "1_3" through "1_6" are such cases. Furthermore, our additional experiment on filter size in Table 2 is a similar case. When we apply convolution on the features of bottom layers, applying larger convolutional filter on an intermediate feature has no remarkable difference than smaller filter. However, as blob passes through several convolution layers, the feature maps from later layers requires larger filters since it contains semantic objects more obviously and estimated semantic object size is larger.

In our model and experiment, collecting the object clues from intermediate layers requires two convolution layers. The filter size of the first convolution layer may vary and empirically chosen. In our experiment, the result inspires us with the suitable convolutional filter size. Larger filter size could be effective in many cases, but it increases time and device requirements. It is the most effective on the intermediate layer marked as "6" in Fig 2. The size of the output feature map from the layers marked as "5" and "6" is 33 × 33. Compared to number "5", the feature map from number "6" gives remarkably better result with bigger filter size than smaller.

## 6. Discussion and Conclusion

Our in-depth research of skip connection in up-to-date deep neural network, inception-v3, reveals the characteristics of each skip connection and the most effective skip connection for semantic segmentation. Besides, it suggests the method of connecting the intermediate features to the uppermost features. Cutting-edge deep neural network consists of numerous layers and has many potential skip connections. Since the recent trend of deep neural network goes deeper, semantic segmentation with neural network would go deeper. In the latest trend of deep network architecture, our analysis will guide us to the best skip connection.

In future, in semantic segmentation task, a recent model of deep neural network needs to be designed for pixel-wise prediction instead of box-wise prediction. Then, it needs to outpace the state-of-the-art performance of semantic segmentation. We plan to train the model from scratch with more training datasets and apply refining technics such as conditional random field or dilated convolution. If we secure pixel-wise prediction architecture, we also plan a task-specific implementation, such as human-pose estimation or pedestrian detection.

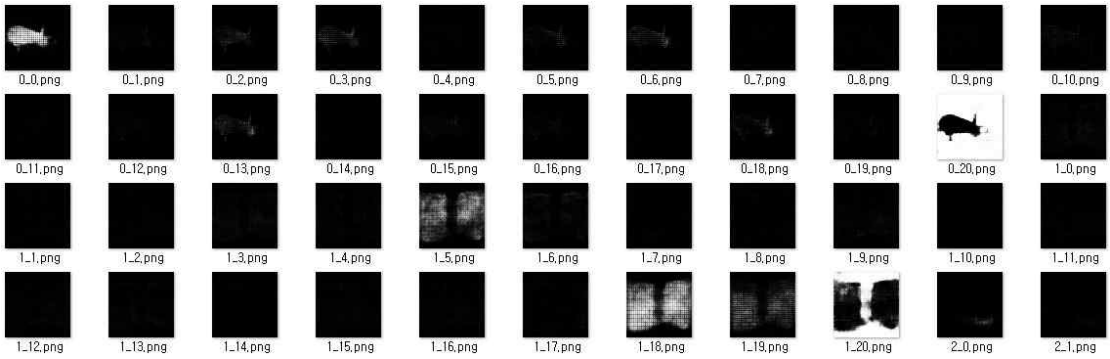

**Fig. 4. Skip connection marked as number 6. The feature map contains well-distinguishable objects and semantic information. Moreover, it tends not to react on false-positive cases. Therefore, the feature maps of negative cases are almost black.**

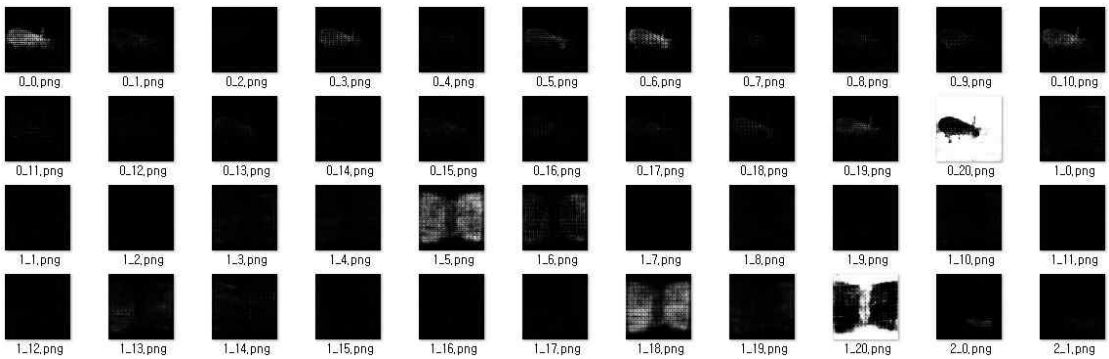

**Fig. 5. Skip connection marked as number 4. Compared to Fig. 4, the feature map contains less semantic information, but it still detects objects and background well.**

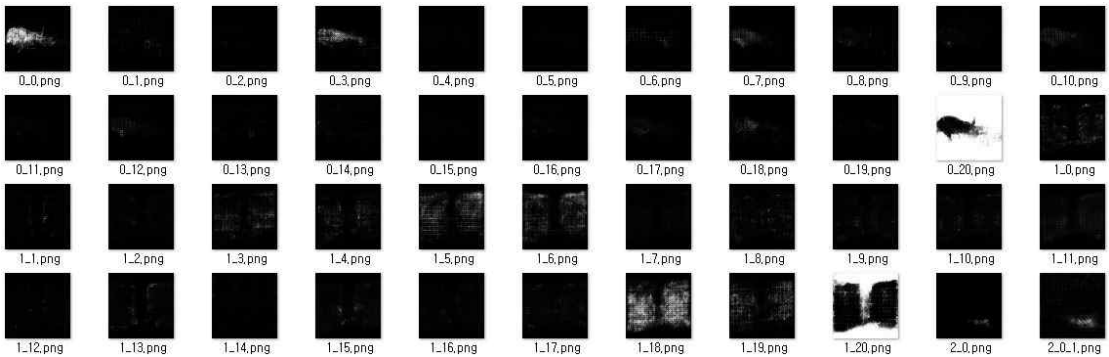

**Fig. 6. Skip connection marked as number2. As this skip connection uses features from a lower layer, it responds to basic components such as edges. Therefore, it reacts on many false classes as well, but still distinguishes background well.**

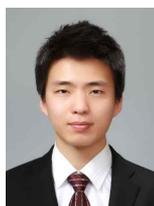

Jonghwa Yim
2013년 한양대학교 컴퓨터공학과 졸업(학사). 2017년 아주대학교 컴퓨터공학과 졸업(석사). 2013년~현재 삼성전자 무선사업부 연구원. 관심분야는 Machine Learning, Computer Vision, Artificial Intelligence, Data Mining

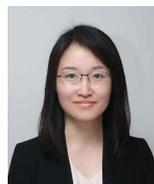

Kyung-Ah Sohn
2000년 서울대학교 수학과 학사. 2003년 서울대학교 컴퓨터공학과 석사. 2011년 Carnegie Mellon Univ. 박사. 현재 아주 대학교 소프트웨어학과 부교수. 관심분야는 기계 학습, 데이터 마이닝, 의생명정보학